%% file: main.tex
\pdfoutput=1
\documentclass[11pt]{article}

\usepackage[preprint]{acl}

\usepackage{times}
\usepackage{latexsym}
\usepackage[T1]{fontenc}
\usepackage[utf8]{inputenc}
\usepackage{microtype}
\usepackage{inconsolata}
\usepackage{graphicx}

\usepackage{booktabs}
\usepackage{amsmath}
\usepackage{amssymb}
\usepackage{amsthm}
\usepackage{array}
\usepackage{algorithm}
\usepackage{algorithmic}
\input{sections/prompts/promptbox}

\graphicspath{{figures/}}

\newtheorem{definition}{Definition}

\newcommand{\traj}{\tau}
\newcommand{\champion}{\mathcal{C}}

\title{The Between-Round Decision in Autonomous Research: An LLM Controller That Reads the Whole Trajectory}

\author{
  Caoxin Wang\textsuperscript{1}\thanks{\ \ Equal contribution.},
  Shengkai Gan\textsuperscript{1}\textcolor{darkblue}{\footnotemark[1]},
  Xiang Hu\textsuperscript{2},
  Lili Pan\textsuperscript{1}\thanks{\ \ Corresponding author.} \\
  \textsuperscript{1}University of Electronic Science and Technology of China, Chengdu, China \\
  \textsuperscript{2}Zhejiang University, Hangzhou, China \\
  \texttt{lilipan@uestc.edu.cn}
}

\begin{document}
\maketitle

\input{sections/abstract}
\input{sections/introduction}
\input{sections/related_work}

\input{sections/method}
\input{sections/evaluation}
\input{sections/conclusion}

\input{sections/limitations}
\input{sections/ethics}

\bibliography{references}

\appendix
\input{sections/appendix}

\end{document}

%% file: sections/abstract.tex
\begin{abstract}
Executing an idea often changes how promising it looks, so the next step in
autonomous research should rest on results, not on proposals. After each round
of experiments, the system must decide whether to refine the best solution or
start a new direction, which solution to build from, and what goal to set. We
call this the between-round decision. Existing systems wrap it in machinery such
as rules, scores, search trees, and schedules; where an LLM makes it, what the
LLM needs to read is left open. We ask whether an LLM can make this decision
from accumulated evidence, without a search structure around it, and how much
of that evidence it needs to read. We present BRED (Between-Round Evidence-based
Decision), an LLM-based controller that maintains the best verified solution so
far (the Champion) and a log of past rounds, and makes one decision per round.
On three widely used research-agent benchmark tasks, BRED improves on the
task baselines by $4.80\%$ in MAE on ETTh1 (to $0.4172$) and by $3.44$ and
$3.43$ points of accuracy on SST-2 (to $95.07\%$) and CIFAR-100 (to
$84.63\%$), ahead of the published results. When the controller does not read
the accumulated evidence, the same loop gains less: with no history it reaches $0.4299$ on ETTh1 and
$83.71\%$ on CIFAR-100, whereas the full controller's MAE is $2.95\%$ lower and
its accuracy $0.92$ points higher.
On a new task, speech-based depression detection under a leakage-free
protocol, it reaches a macro-F1 of $0.6164$ on seeds held out from selection,
$5.5$ points above its baseline and $4.0$ above the strongest published system.
Code is available at \url{https://github.com/learninginvision/BRED}.
\end{abstract}

%% file: sections/introduction.tex
\section{Introduction}
\label{sec:intro}

\begin{figure*}[t]
    \centering
    \includegraphics[width=\textwidth]{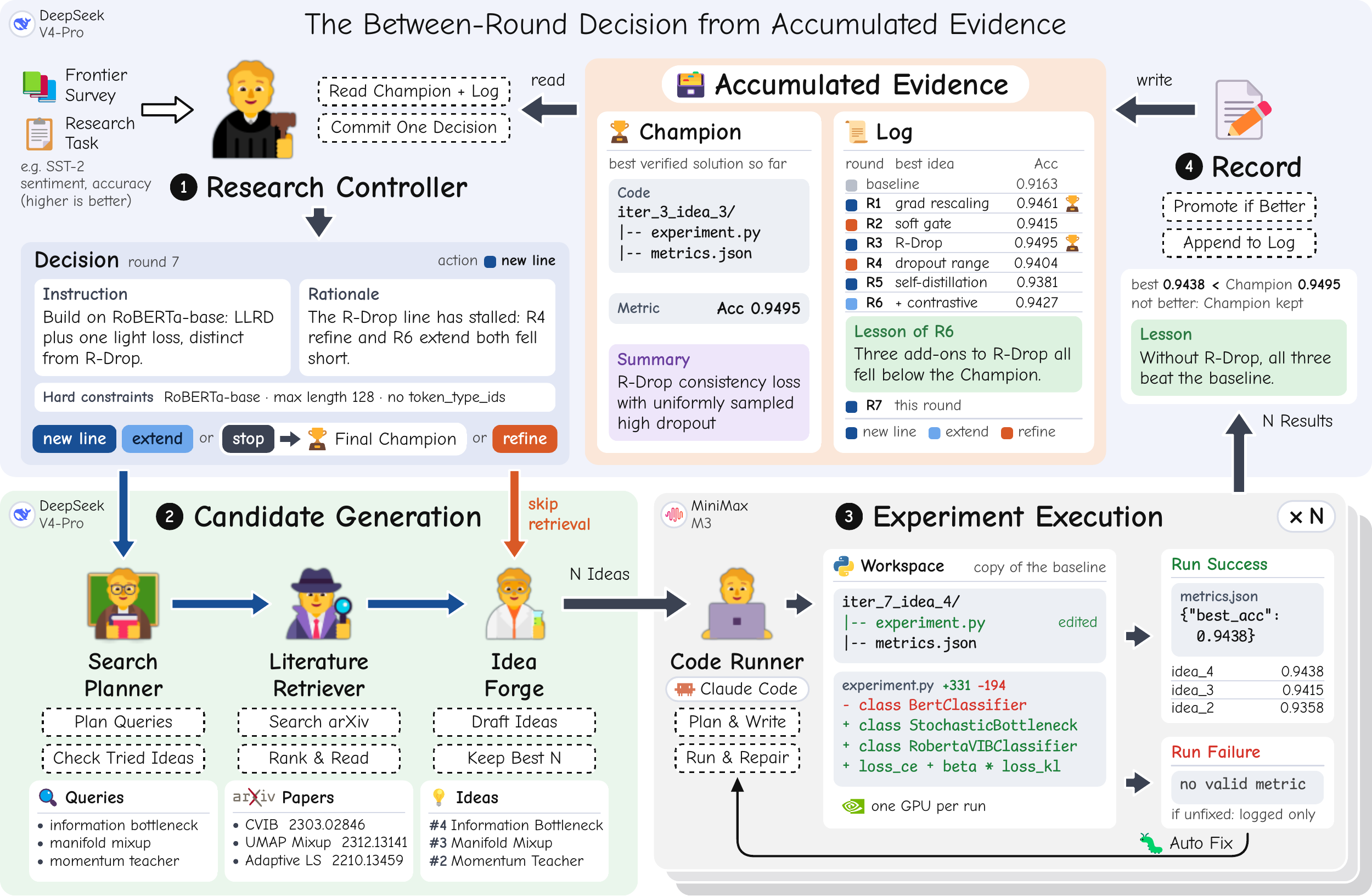}
    \caption{Overview of the loop, shown on round~7 of the AutoClsSST run.
    (1)~The research controller reads the accumulated evidence, namely the Champion and the log,
    together with the task and a survey of its published lines, and commits one decision: one of
    four actions (start a new line from the baseline, extend the Champion, refine it, or stop), an
    instruction, a rationale, and hard constraints. (2)~For a new line or an extension, candidate
    generation plans queries, retrieves papers, and drafts ideas; refinement skips retrieval.
    (3)~Each of the $N$ ideas is implemented and run in its own copy of the starting workspace,
    and failures are repaired. (4)~Record promotes the round's best candidate only if it beats
    the Champion, and appends the round and its lesson to the log. No weights are updated.}
    \label{fig:method}
\end{figure*}

An autonomous research system proposes ideas, implements and runs them, and
then has to decide what to do next. This decision has to rest on execution
results, because what an idea is worth is not visible until it has been run.
When experts executed LLM-generated research ideas, the ideas' review scores
dropped significantly more than those of human-written ideas on every metric,
and on many metrics the ranking between the two flipped~\citep{si2025ideationexecution}.
Each round therefore ends with a decision based on results: whether to refine
the best solution so far or start in a different direction, which earlier
solution to build from, and what goal to give the next round. We call this the
\emph{between-round decision}.

Existing systems place this decision in different parts of the loop. Some fix
it with a hard-coded policy, a search rule, or a schedule. AIDE applies a
hard-coded policy that drafts, debugs, or improves the best
solution~\citep{jiang2025aide}. SELA chooses among candidate insights with a
UCT variant over scalar rewards~\citep{chi2024sela}, and PrimeScientist samples
nodes with a score rule that concentrates as the budget is
spent~\citep{yu2026primescientist}. \citet{si2026executiongrounded} anneal the
share of exploitation over epochs on a schedule fixed in advance.
Others work on the ideas themselves. Dolphin feeds results into the next
round's idea-generation prompt and filters out ideas that resemble ones that
did not help~\citep{yuan2025dolphin}, and InternAgent ranks and evolves ideas
with agent scores before they are executed~\citep{novelseek2025}. More recent
systems hand the choice to an LLM. AI Scientist v2 has an LLM pick the best
node among those that ran~\citep{yamada2025aiscientistv2}. ARTS and
SwarmResearch have an LLM choose which node to extend and whether to explore or
refine~\citep{juneja2026arts,virk2026swarmresearch}. AlphaLab's LLM strategist
plans the next experiments, while a budget counter written into its prompt sets
how selective it should be~\citep{hogan2026alphalab}.

Most of these designs build machinery around the decision: a tree, a score, a schedule, a
filter. Where an LLM makes the decision, what it needs to see is left open. The ablations we
are aware of switch a memory or a playbook on and off~\citep{juneja2026arts,hogan2026alphalab}, and a
comparison of an LLM scheduler with a bandit on a two-way choice found the bandit
better~\citep{li2025rdagentquant}. We ask the question directly: can an LLM make the
between-round decision from accumulated evidence alone, without a search structure
around it, and how much of that evidence does it need?

To study this, we build a between-round controller (Figure~\ref{fig:method}) that keeps two
objects. The \emph{Champion} is the best verified solution, stored with its
code, its metric, and a short summary of its mechanism. The \emph{log} has one
entry per round, holding the decision taken, the results of the candidates, and
a short factual lesson. Each round, one LLM call reads both objects and returns
a decision: an action, which also fixes the solution to start from, a
goal-level instruction, optional hard constraints, and a rationale. The action
is one of four: start a new line from the
clean baseline; extend the Champion with a new mechanism, using retrieved
literature; refine the Champion without adding a mechanism; or stop. A few
guardrails in code remain, such as reserving the last 30\% of rounds for
refining or stopping (Section~\ref{sec:decide}); within these limits, the
choices are the LLM's. Retrieval, idea generation, and coding use ordinary components, and no
weights are updated.

In an ablation study, we vary only what the decision step is shown and keep
everything else fixed. There are three settings: no history, where the
controller sees neither the Champion nor the log and every round restarts from
the baseline; 1-round history, where only the last round is shown; and full history, where the controller reads the Champion together
with the full log. We run all three on time-series forecasting and CIFAR-100
image classification and report the best verified result of each. Performance
improves as the decision step is shown more of the history. With no history,
the loop reaches an MAE of $0.4299$ on forecasting and $83.71\%$ accuracy on
classification. Showing the last round lowers the MAE to $0.4244$ and raises
the accuracy to $84.01\%$, and the full controller reaches $0.4172$ and
$84.63\%$ (Section~\ref{sec:ablation}).

In summary, we make three contributions.
\begin{enumerate}
  \item \textbf{We propose BRED}, an LLM controller for the between-round
    decision in autonomous research. It keeps only a Champion (the best
    verified solution) and a log of past rounds, makes one decision per
    round, and runs unchanged across four tasks.
  \item \textbf{We study how much accumulated evidence the controller needs.} In three
    context settings on two tasks (no history, the last round only, and the
    full history), performance improves monotonically as the controller reads more
    of the accumulated evidence.
  \item \textbf{We improve on published systems.} With the full controller and
    the same prompts on all four tasks, BRED reaches an MAE of $0.4172$ on
    ETTh1, $95.07\%$ accuracy on SST-2, and $84.63\%$ on CIFAR-100, against
    published results of $0.4331$, $93.50\%$, and $83.30\%$, respectively; each
    figure is the best final Champion, following the benchmark's reporting
    convention. On speech-based depression detection under a leakage-free
    protocol, BRED reaches a macro-F1 of $0.6164 \pm 0.0166$ on five seeds held
    out from selection, against $0.576$ for the strongest published
    system~\citep{li2026haren}.
\end{enumerate}

%% file: sections/related_work.tex
\section{Related Work}
\label{sec:related}

\paragraph{Autonomous research systems.}
AI-scientist systems automate idea generation, experiment execution, and paper
writing~\citep{lu2024aiscientist,yamada2025aiscientistv2}.  Dolphin~\citep{yuan2025dolphin} and
InternAgent~\citep{novelseek2025} close the loop over ideation, coding, and
feedback on widely used research-agent benchmark tasks.
Other work closes the
loop in chemistry~\citep{boiko2023autonomous}, reasons over knowledge graphs for
discovery~\citep{ghafarollahi2024sciagents}, or evaluates data-driven scientific
agents~\citep{wang2024scienceagentbench}.
Long-horizon evaluations find that
such agents are impatient, over-trust weak hypotheses, and allocate a finite
experiment budget poorly~\citep{garikaparthi2026researchgym}. 

\paragraph{Where the between-round decision sits.}
Related work falls into three groups. First, search-based systems obtain the
balance between refining and exploring from a search rule: AIDE from a
hard-coded policy over a tree of solutions~\citep{jiang2025aide}, SELA from a
UCT term~\citep{chi2024sela}, PrimeScientist from a score
rule~\citep{yu2026primescientist}, and AI Scientist v2 from node selection in an
experiment tree~\citep{yamada2025aiscientistv2}. Their state is a tree of scored
nodes; ours is the Champion and a linear log, and each move comes with a stated
reason. Second, \citet{si2026executiongrounded} ground idea selection in
executed results through evolutionary search, with the share of exploitation
annealed on a schedule fixed before the run and a signal drawn from hundreds of
parallel executions per epoch. Our controller instead decides each round from
what the run has found so far, with three to five candidates per round. The same
study finds that GRPO on execution reward raises the average but not the
maximum reward, with the policy collapsing onto trivial edits, which is
consistent with leaving our controller untrained. Third, ARTS, SwarmResearch,
and AlphaLab, like us, hand the decision to an
LLM~\citep{juneja2026arts,virk2026swarmresearch,hogan2026alphalab}. Their
ablations switch a memory or a playbook on and off, whereas we vary, in three
steps, how much of the accumulated evidence the controller reads.

\paragraph{History in language agents.}
Sequential decision-making can condition on history through recurrent policies
or episodic memory~\citep{hausknecht2015deep,wang2016learning}, usually at the
cost of training. Language agents keep history in natural language instead.
Reflexion stores verbal reflections on failed trials~\citep{shinn2023reflexion},
ExpeL distils past trajectories into reusable insights~\citep{zhao2024expel},
and Voyager accumulates verified skills over a lifelong
run~\citep{wang2023voyager}. These systems use history to condition the next
\emph{attempt} at a task. We use it to condition the between-round
\emph{decision}: which line to pursue, what to build on, and when to stop. Our
log also keeps facts apart from interpretation. Its lessons record only what was
measured, and the reading of the evidence is left to \textsc{Decide}.

\paragraph{Retrieval and the explore--exploit tradeoff.}
Retrieval-augmented generation conditions an LLM on documents fetched from an
external index~\citep{lewis2020rag}, and later work lets the model decide when to
retrieve: FLARE retrieves when an upcoming sentence contains low-confidence
tokens~\citep{jiang2023flare}, Self-RAG learns special tokens that trigger
retrieval and critique its own output~\citep{asai2023selfrag}, and ReAct
interleaves reasoning with tool calls such as search~\citep{yao2022react}.
Choosing between gathering new information and using what is already known is the
explore--exploit tradeoff studied in bandits and reinforcement
learning~\citep{sutton2018reinforcement}. Unlike these works, which make the
choice within a single answer or among a fixed set of arms, our controller makes
it once per research round, between retrieving papers to open or extend a line
and refining the Champion without retrieval.

%% file: sections/method.tex
\section{BRED Pipeline}
\label{sec:method}

BRED runs four steps per round, \textsc{Decide}, \textsc{Generate},
\textsc{Run}, and \textsc{Record}, over two persistent objects
(Figure~\ref{fig:method}, Algorithm~\ref{alg:loop}). \textsc{Decide} chooses
what the round does; the other three steps carry it out.

\subsection{Persistent State: the Champion and the Log}
\label{sec:problem}

A task comes with a baseline workspace, a frozen
evaluation protocol, and a budget of $T$ rounds, and the verified baseline
metric is the floor $\mathcal{B}$. Each round produces candidates, which are
implemented ideas. Each candidate has an identifier, a workspace, and, when its
execution succeeds, a metric. After a candidate runs, an LLM reads its actual
code and records a mechanism signature of at most twelve words, its key
configuration, and a note on its likely bottleneck. The loop thus keeps what
was built rather than what was proposed.

\begin{definition}[Champion]
The Champion after round $t$ is the best verified candidate found in rounds
$1,\ldots,t$, recorded as $\champion_t=(i^*_t,w^*_t,m^*_t,s^*_t,r^*_t)$, where $i^*_t$ is its identifier, $w^*_t$ its workspace, $m^*_t$ its metric,
$s^*_t$ its mechanism signature, and $r^*_t\le t$ the round that produced it.
Before any candidate strictly improves on the floor $\mathcal{B}$ (in the
task's metric direction), $\champion_t=\varnothing$. Afterwards, $\champion_t$
changes only when a later candidate strictly improves on $m^*_t$, so $m^*_t$
never regresses.
\end{definition}

\begin{definition}[Log]
The log entry of round $t$ is the tuple $e_t=(d_t,R_t,b_t,u_t)$, where $d_t$ is the decision taken, $R_t$ the results of all candidates (a
metric for each success, a diagnosis for each failure), $b_t$ the best metric
of the round, and $u_t$ a short factual lesson. The log after round $t$ is the
ordered sequence $\traj_{1:t}=(e_1,\ldots,e_t)$, with $\traj_{1:0}=()$.
\end{definition}

The lesson states only what the round's data verify: what was tried, with which settings, and
how it scored. It may not attribute an outcome to a mechanism, so that one noisy measurement
does not become a causal story that later rounds inherit; interpreting the evidence is left
to Decide.

\begin{algorithm}[t]
\caption{The BRED Iteration Loop.}
\label{alg:loop}
\begin{algorithmic}[1]
\STATE $\champion_0 \leftarrow \varnothing$,\ \ $\traj_{1:0} \leftarrow ()$,\ \ floor $\mathcal{B}$
\FOR{$t=1,2,\ldots,T$}
  \STATE $d_t \leftarrow \mathrm{Decide}(\mathcal{B},\champion_{t-1},\traj_{1:t-1},T-t+1)$
  \IF{$d_t$ is \texttt{stop}}
    \STATE \textbf{break}
  \ENDIF
  \STATE $C_t \leftarrow \mathrm{Generate}(d_t,\champion_{t-1})$
  \STATE $R_t \leftarrow \mathrm{Run}(C_t)$
  \STATE $(e_t,\champion_t) \leftarrow \mathrm{Record}(d_t,R_t,\champion_{t-1},\mathcal{B})$
  \STATE $\traj_{1:t} \leftarrow \traj_{1:t-1} \oplus e_t$
\ENDFOR
\STATE \textbf{return} the final Champion and the log
\end{algorithmic}
\end{algorithm}

\subsection{Decision: One LLM Call per Round}
\label{sec:decide}

Decide is a single LLM call per round; no policy is learned. The prompt holds the task goal
and metric direction, the floor, the Champion with its signature and bottleneck note, the log
(one line for every round, plus the decisions, rationales, and all candidates of the last five
rounds in full), the remaining budget, and the number of consecutive rounds in which
extending the Champion has not improved it. It also holds a survey of the task's published
research lines and their reported results, which the controller uses to set targets; the
survey is the same in every round and in every setting of Section~\ref{sec:ablation}.
Appendix~\ref{app:decide-prompt} gives the prompt and the output schema.

The model first assesses the situation and then returns a decision $d_t$ with four parts: an
action, an instruction, hard constraints, and a rationale. The \emph{action} is one of four: start a new line from the clean baseline
(\texttt{explore}/\texttt{baseline}); extend the Champion with a new mechanism built on its
code (\texttt{explore}/\texttt{Champion}); refine the Champion's code without a new mechanism
(\texttt{exploit}); or \texttt{stop} and deliver the Champion. The \emph{instruction} states
the round's goal in at most 40 words, naming the bottleneck to attack and the level to aim
for rather than a recipe. In exploration rounds, up to three \emph{hard constraints} reach every candidate verbatim,
and whether they were followed is recorded in the log. The \emph{rationale} cites the
evidence. The action drives control flow and the instruction carries the scientific content:
``the Champion is limited by cross-channel modelling; aim for the $0.42$ level'' is an
instruction, not a new action type.

After parsing, guardrails in code may override the action: rounds start from the baseline until
a first Champion exists, and the last 30\% of rounds allow only refining or stopping. Before
that closing phase, a further extension after three consecutive extensions without progress
becomes a new line (the prompt suggests switching after two), and a stop is refused if the
Champion improved in either of the last two rounds. An overridden action gets a fixed
instruction and no hard constraints.

\subsection{Round Execution: Candidates, Results, and the Champion Update}
\label{RoundExe}

\textsc{Generate} turns the decision into $N$ candidates, with $N$ fixed per task, using the
three agents of Figure~\ref{fig:method} (Appendix~\ref{app:generate}). For both exploration
actions, the Search Planner turns the instruction, the survey, and a ledger of tried directions
into six search plans; for each plan, the Literature Retriever reads in full the three best of
up to 30 arXiv papers; and the Idea Forge drafts one idea per plan on the starting code, which
is the Champion's code when the Champion is extended. The $N$ most feasible ideas that pass a
feasibility and a grounding check are run, or the $N$ most feasible of all if fewer pass.
Refinement skips retrieval: the Idea Forge proposes five incremental changes to the Champion's
code, selected the same way. \textsc{Run} hands each candidate to a coding agent that copies the
starting workspace, implements the change, executes it under the frozen protocol, and repairs
failures, returning a metric or a diagnosis; candidates run in parallel. \textsc{Record}
discards diverged results, and the round's best candidate replaces the Champion only if it
strictly improves on it (on $\mathcal{B}$ before a Champion exists). The Champion therefore
never regresses, every refinement starts from the best code so far, and a failed round costs
nothing already won. Record then appends the round to the log. Appendix~\ref{app:design} lists
the structures this design leaves out.

\subsection{Instantiating BRED}
\label{sec:implementation-main}

The BRED pipeline is orchestrated with LangGraph. \textsc{Decide} is one \texttt{deepseek-v4-pro} call per
round, so the controller uses no local GPU time, and Generate uses ordinary planning, retrieval, and
idea-generation components whose LLM calls go to the same model. Run delegates each candidate to a Claude Code
coding agent backed by \texttt{MiniMax-M3}, which writes, executes, and repairs the
candidate's code in its own workspace; the loop never writes experiment code itself. Any
execution stack that turns an instruction into a code change in a given workspace, reports a
metric, and keeps the Champion's workspace can host the controller
(Appendix~\ref{app:implementation}).

%% file: sections/evaluation.tex
\section{Evaluation}
\label{sec:evaluation}

\begin{table*}[t]
\centering
\small
\setlength{\tabcolsep}{5pt}
\begin{tabular}{l l r r r r r r}
\toprule
 & & & & \multicolumn{2}{c}{InternAgent} & & $\Delta$ vs.\ \\
\cmidrule(lr){5-6}
Task (dataset) & Metric & Baseline & Dolphin & published & re-run$^\dagger$ & Ours & best prior \\
\midrule
AutoTSF (ETTh1)        & MAE $\downarrow$      & 0.4382 & 0.4627 & 0.4331 & 0.4245 & $\mathbf{0.4172}$ & $-1.72\%$ \\
AutoClsSST (SST-2)     & Acc.\ $\uparrow$      & 91.63  & 92.50  & 93.50  & 92.78  & $\mathbf{95.07}$  & $+1.57$ pp \\
AutoCls2D (CIFAR-100)  & Acc.\ $\uparrow$      & 81.20  & 82.00  & 83.30  & 83.44  & $\mathbf{84.63}$  & $+1.19$ pp \\
AutoDepDAIC (DAIC-WOZ) & Macro-F1 $\uparrow$   & 56.10  & ---    & ---    & ---    & $\mathbf{61.64}$  & $+4.04$ pp \\
\bottomrule
\end{tabular}
\caption{Main results. Dolphin and published InternAgent numbers are taken verbatim from
    Tables~1--2 of \citet{novelseek2025}, reported there and here as maximum performance across
    runs; re-run$^\dagger$ is InternAgent's released code run with our models
    (Appendix~\ref{app:internagent-rerun}). The first three tasks share that benchmark's
    datasets, baselines, and metrics. AutoDepDAIC lies outside it and is compared against
    HAREN-CTC~\citep{li2026haren} ($57.6$), the strongest published system under the same
    leakage-free protocol; our entry there is measured on five seeds that took no part in
    selection. AutoTSF, AutoClsSST, and AutoDepDAIC baselines are our runs' verified floors.
    The last column compares our result with the strongest prior entry in the row (relative
    reduction for MAE, gap in points otherwise).}
\label{tab:main-results}
\end{table*}

We evaluate the controller on two questions. First, does a loop driven by it reach results
on par with or better than published autonomous-research systems on the same tasks? Second, how much of the
accumulated evidence does the controller need to read? For the first we adopt the task
protocol of the InternAgent benchmark~\citep{novelseek2025}, with the same datasets,
baselines, and primary metrics, and re-run InternAgent's released code with our models
(Appendix~\ref{app:internagent-rerun}). For the second we vary what the \textsc{Decide} step is shown
(Section~\ref{sec:ablation}). Throughout, the reported quantity is the \emph{best verified
metric}: the final Champion measured under the task's own evaluation protocol, in the way
the benchmark reports maxima.

\paragraph{Setup.}
We evaluate on four tasks. Three are widely used research-agent benchmark
tasks: \emph{AutoTSF} (ETTh1 time-series forecasting, MAE, lower is better),
\emph{AutoClsSST} (SST-2 sentiment classification, accuracy), and
\emph{AutoCls2D} (CIFAR-100 image classification, top-1 accuracy). The fourth,
\emph{AutoDepDAIC} (speech-based depression detection on DAIC-WOZ, macro-F1),
lies outside the benchmark and is compared against the published state of the
art for that task. All four tasks run the same controller and prompt templates
with no task-specific control logic; only the task configuration and the
execution environment differ.

In every run, \textsc{Decide}, retrieval, and idea generation use
\texttt{deepseek-\allowbreak v4-\allowbreak pro} at temperature $0.7$, and
\textsc{Run} uses the Claude Code agent backed by \texttt{MiniMax-M3}. Outside the ablation, the
controller reads the Champion and the full log in every round. Experiments
execute on five NVIDIA RTX 2080 Ti GPUs (11\,GB each). On AutoDepDAIC, the
evaluation repeats over five seeds and five folds, as its protocol requires,
and the headline number is measured on seeds held out from selection.
Appendix~\ref{app:trajectories} gives the per-round record of the four main
runs.

\subsection{Comparison with Published Systems}
\label{sec:main-results}

\paragraph{AutoTSF.}
AutoTSF shares the benchmark's ETTh1 dataset, DLinear baseline, MAE metric, and
average over prediction lengths $\{96,192,336,720\}$. The run has a ten-round
budget and three candidates per round (Table~\ref{tab:main-run}).

\emph{Trajectory.} In R1--R3, the controller starts from the baseline with
three strong lines from the task survey, ModernTCN-, PatchTST-, and FITS-style
designs, and none of their candidates beats the floor. At R4, its rationale
reviews all three rounds, concludes that these lines do not beat the baseline
in this setting, and sets a new goal, cross-channel structure. That round
produces the first Champion ($0.4226$). From R5 to R7, the controller extends
the Champion with one new mechanism per round, each aimed at the bottleneck
read from the previous Champion, and the Champion improves in every round, to
$0.4172$. After a closing refinement at R8, the controller stops at R9 with
budget left. Its rationale records that the gains have shrunk to zero over five
rounds and that the Champion already surpasses the strongest line in its
survey.

\emph{Comparison.} The final $\mathbf{0.4172}$ is $4.80\%$ below the DLinear
baseline, $3.67\%$ below InternAgent's published result, and $9.83\%$ below
Dolphin, which is itself $5.59\%$ \emph{worse} than the baseline it started
from. Re-run with our models, InternAgent improves to $0.4245$, ahead of its
own published figure, and our controller still leads it by $1.72\%$
(Section~\ref{sec:same-model}).

\begin{figure*}[t]
    \centering
    \includegraphics[width=\textwidth]{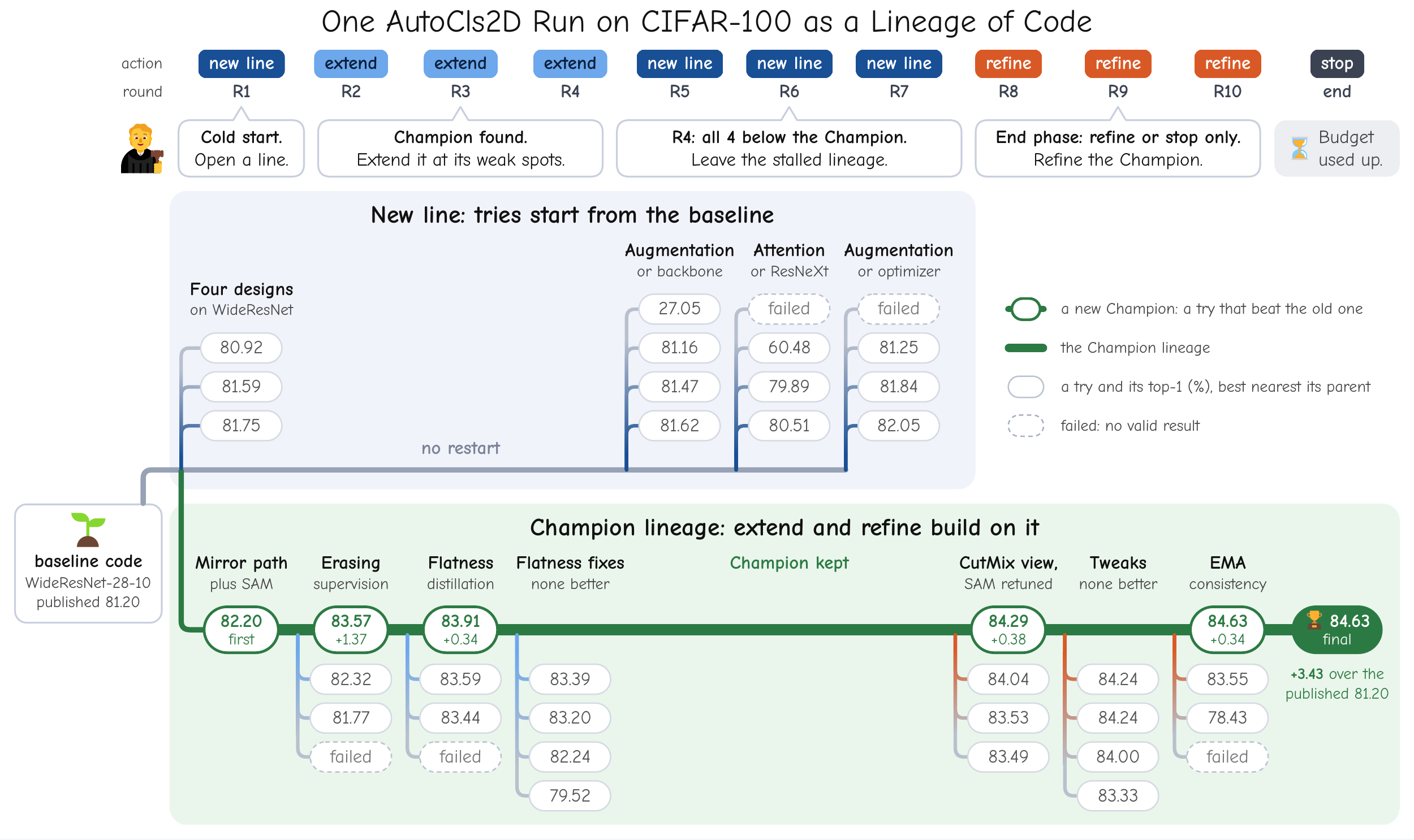}
    \caption{The AutoCls2D run as a lineage of code. The top row gives each round's action and
    the controller's decision. Each candidate hangs from the code it starts from: new lines
    (R1, R5--R7) from the baseline, in the upper band, and extensions and refinements from the
    Champion, below the green Champion lineage, whose nodes mark each new Champion and its
    gain. Pills give each candidate's top-1 accuracy (\%); dashed pills mark candidates
    without a valid result. R1--R3 raise the Champion to $83.91\%$. After R4 fails to extend
    it, the controller leaves the stalled lineage and opens three new lines from the baseline
    (R5--R7); none beats the Champion, which is kept unchanged. R8--R10 fall in the last 30\%
    of rounds, reserved for refining or stopping, and two refinements lift the Champion to
    $84.63\%$, $3.43$ points above the published WideResNet baseline ($81.20\%$). The run ends
    when its ten-round budget is used up. Round labels are read back from the logged ideas and
    each candidate's code.}
    \label{fig:case}
\end{figure*}

\paragraph{AutoClsSST.}
With three candidates per round, the controller follows its survey to a
RoBERTa-base encoder in R1 ($94.61\%$), already above InternAgent's published
result ($93.50\%$). In R3 it starts a new line with consistency-regularised
fine-tuning, which becomes the Champion ($94.95\%$). Refinements right after a
new Champion (R2, R4) and further exploration (R5--R7) do not beat it, so the
Champion stays intact until the closing refinement at R8 raises it to
$\mathbf{95.07\%}$. At R10, the controller stops, noting that the Champion
already exceeds the ceiling its survey gives for this encoder. The final
Champion is $1.57$ points above InternAgent's published result, $2.57$ above
Dolphin, and $2.29$ above InternAgent re-run with our models ($92.78\%$).

\paragraph{AutoCls2D.}
Figure~\ref{fig:case} traces the run, with four candidates per round. R1--R3
build one lineage: a WideResNet-28-10 with an auxiliary mirror pathway and
curvature-aware SAM, and the Champion improves in each round, to $83.91\%$. When
extending it fails at R4, the controller leaves this lineage of its own accord
and opens three new lines from the baseline (R5--R7). None of them beats the
Champion, so it stays intact. In the closing rounds, the controller returns to
the Champion with targeted refinements, the last of which adds an EMA
consistency target on the auxiliary pathway and lifts it to $\mathbf{84.63\%}$.
This is $1.33$ points above InternAgent's published result, $2.63$ above
Dolphin, and $1.19$ above InternAgent re-run with our models ($83.44\%$).

For the baseline, we use the published WideResNet result ($81.20\%$). The
in-repository baseline is a three-epoch smoke configuration ($49.23\%$) that
serves only as the floor before the first Champion.

\paragraph{AutoDepDAIC.}
AutoDepDAIC is a new task outside the existing benchmark: speech-based depression
detection on DAIC-WOZ. The controller runs on it unchanged, with the same
prompts and no task-specific control logic, and we compare the result with the
published systems for this task.

\emph{Protocol.} DAIC-WOZ has a documented history of inflated results, so we
adopt the leakage-free \emph{generalization evaluation} of
HAREN-CTC~\citep{li2026haren}: 188 subjects pooled, subject-stratified 5-fold
cross-validation over five seeds, with every training, hyper-parameter, and
threshold choice confined to each fold's training part
(Appendix~\ref{app:depdaic}). Under this protocol, that system's own macro-F1
falls from $0.81$ to $0.576$. The protocol also awards nothing to a model
without signal: the all-positive predictor scores only $\approx 0.23$, and our
port of DepAudioNet~\citep{ma2016depaudionet}, the most-cited audio baseline on
this corpus, performs at chance (macro-F1 $0.474$, AUC $0.467$).

\emph{Run.} With five candidates per round, the Champion reaches $0.5994$ at R2
and survives four rounds that do not improve it, two of which return no metric.
From R7, the controller refines it to $0.6019$, and at R10 it stops, recording
that three refinement rounds have each moved the metric by less than $0.003$
(Appendix~\ref{app:depdaic} describes the discovered model).

\emph{Result.} Because $0.6019$ is the maximum over 44 candidates scored on the
same folds, we re-ran the unchanged Champion code on five fresh seeds that took
no part in any selection. It reaches a macro-F1 of $\mathbf{0.6164 \pm 0.0166}$
and an AUC of $\mathbf{0.6902 \pm 0.0151}$, with every seed at or above $0.5987$
macro-F1. This fresh-seed estimate is our headline number. It exceeds
HAREN-CTC, which is closed-source and the strongest published system under this
protocol ($0.576$ macro-F1, $0.599$ AUC), by $4.0$ points of macro-F1 and $9.1$
points of AUC, and it exceeds our own $0.5610$ floor by $5.5$ points of macro-F1.

\subsection{Unproductive Rounds}

Every run contains rounds that do not improve on the best result so far:
AutoTSF's R1--R3, six of AutoClsSST's nine rounds, AutoCls2D's R4--R7 and R9,
and AutoDepDAIC's R3--R6 and R9. These rounds cost the loop nothing it has already
won, because the Champion changes only on a verified improvement. Each run also
recovers from them, by setting a new goal on AutoTSF and by refining the
Champion on the other three. We show in
Section~\ref{sec:ablation} what the loop reaches when the decision step reads
none of this evidence.

\subsection{Same-Model Control}
\label{sec:same-model}

Our execution stack and models differ from those behind the published figures,
so we also re-ran InternAgent's released code with the models and the coding
agent that our controller uses. It falls behind our controller on all three
tasks (Table~\ref{tab:main-results}), so the gap is not a matter of the backbone.

The two loops differ in what they iterate on. InternAgent follows its original
protocol: it iterates on ideas before any execution and then runs the three
top-ranked ones. Our loop iterates on execution, running up to ten rounds, each
decided from the results of the earlier ones.

The ablation provides a second check. With the same budget but no history, our
loop reaches an MAE of $0.4299$ on AutoTSF, behind the re-run InternAgent
($0.4245$), while the full controller is ahead ($0.4172$). On AutoTSF, the
margin therefore appears only when the controller reads accumulated evidence,
not from running more experiments. On AutoCls2D, our loop is ahead even without
history, and reading the full context widens the margin.

\subsection{Ablation: How Much Evidence Does the Controller Need?}
\label{sec:ablation}

\begin{table}[t]
\centering
\small
\setlength{\tabcolsep}{4pt}
\begin{tabular}{@{}l c c@{}}
\toprule
Decide reads & AutoTSF & AutoCls2D \\
 & MAE $\downarrow$ & Acc.\ $\uparrow$ \\
\midrule
No history             & $0.4299$          & $83.71$ \\
1-round history        & $0.4244$          & $84.01$ \\
Full history (ours)    & $\mathbf{0.4172}$ & $\mathbf{84.63}$ \\
\bottomrule
\end{tabular}
\caption{Best verified result of each context setting. Baselines: AutoTSF $0.4382$;
    AutoCls2D $81.20\%$~\citep{novelseek2025}.}
\label{tab:ablation-ladder}
\end{table}

To see how much of the accumulated evidence the controller needs, we vary only
what the \textsc{Decide} step is shown and keep everything else fixed: the LLM
and its temperature, the prompt skeleton and output schema, the
\textsc{Generate}, \textsc{Run}, and \textsc{Record} layers, and the budget (ten
rounds; three candidates per round on AutoTSF, four on AutoCls2D). There are
three settings:
\begin{itemize}
  \item \emph{no history}: the decision step sees neither the Champion nor the
    log, so every round starts from the baseline;
  \item \emph{1-round history}: it sees only the last round, not the Champion;
  \item \emph{full history}: it sees the Champion and the full log. This is
    our controller.
\end{itemize}
\textsc{Generate} reads the full log in every setting (Appendix~\ref{app:generate}).

\paragraph{Result.}
On both tasks, the controller is weakest with no history, and performance
improves monotonically as it is shown more context
(Table~\ref{tab:ablation-ladder}): even 1-round history improves over no
history, and the full controller finishes best on both tasks.

\paragraph{What changes in the search.}
The AutoTSF runs show how. Under no history, the best result, $0.4299$ ($1.9\%$
better than the baseline), is reached at R4 and does not improve in the
remaining six rounds, and some directions are repeated across rounds. The full
controller (Section~\ref{sec:main-results}) also spends R1--R3 without beating
the baseline, but then builds on the line it finds at R4 until it stops at R9.
Appendix~\ref{app:ablation-rounds} compares when each setting first beats the
baseline.

%% file: sections/conclusion.tex
\section{Conclusion}
\label{sec:conclusion}

We asked two questions: can an LLM make the between-round decision of an
autonomous research loop from accumulated evidence, and how much of that
evidence does it need to read? To answer them, we built BRED, a controller that
keeps a Champion and a log of past rounds and makes one decision per round,
with no search structure around it.

BRED outperforms the published systems on three benchmark tasks, and it
outperforms InternAgent re-run with the same models. On speech-based depression
detection under a leakage-free protocol, it exceeds the strongest published
system. When we vary what the decision step reads, the loop is weakest with no
history, and its performance improves as it reads more of the accumulated
evidence. The between-round decision can therefore be made with less machinery
than existing systems build around it, provided the controller sees the
accumulated evidence. The monotonic improvement along the context ladder
suggests that richer trajectory representations, such as structured failure
summaries or explicit mechanism lineages, may yield further gains; we leave this
to future work.

%% file: sections/limitations.tex
\section*{Limitations}

All our runs use one controller LLM, and we have not measured how the quality of its
decisions changes with other models. The ablation covers two tasks; it shows
that the gain grows with context, but does not decompose which component of the accumulated
evidence contributes most. Dolphin's numbers are taken from \citet{novelseek2025}
and were obtained with other models (GPT-4o for ideation, Claude-3.7-Sonnet for coding) and
other hardware; for InternAgent we add a re-run with our models
(Appendix~\ref{app:internagent-rerun}). We report three of the benchmark's twelve tasks,
covering three modalities, and one task outside it. Several of the remaining tasks are
project-level, spanning repository-scale codebases that the benchmark notes one of the
compared systems cannot modify, and we did not attempt them. Finally, the lessons in the log are written by an LLM; we
restrict them to facts, but a wrong lesson would still reach later decisions.

%% file: sections/ethics.tex
\section*{Ethical Considerations}

Our controller lowers the cost of running a closed research loop without human supervision.
The benefit is that an inspectable controller can carry a verified solution forward over many
rounds, and that every decision it makes is recorded in natural language and can be audited
after the fact. The risks follow from the same property. A loop that optimises a single
reported metric will exploit whatever the metric fails to constrain, which is why the
evaluation protocol is frozen in the execution layer rather than left to the generated code,
and why one of our tasks is evaluated under a leakage-free protocol whose canonical baseline
measures at chance. Systems of this kind also make it cheap to produce large volumes of
plausible-looking empirical claims; the log is the mitigation we rely on, since it keeps each
claim attached to the run that produced it.

AutoDepDAIC uses DAIC-WOZ, a corpus of clinical interviews distributed for academic research
under an end-user license agreement. We use it under that agreement, report only aggregate
metrics, and release no audio, transcripts, or per-subject predictions. A model trained on
188 subjects that reaches macro-F1 near $0.6$ is a research artifact for comparing
controllers, not a screening instrument, and should not be used to diagnose depression.

\paragraph{Use of AI assistants.}
We used AI assistants (Claude) to help develop the code, draft and revise the text, and
prepare the figures of this paper. The authors reviewed all of it and take full responsibility
for the content.

%% file: sections/appendix.tex
\section{Design Choices}
\label{app:design}

We add structure only where it changes execution. The ablation of
Section~\ref{sec:ablation} shows that the two objects we keep matter: when \textsc{Decide}
reads neither the Champion nor the log, the loop is weakest on both tasks. It does not test
whether the structures in Table~\ref{tab:removed} would add further gains; we leave them out
for simplicity, because the Champion and the log already answer the three questions the
controller faces each round: is the loop still making progress, what has been tried and
learned, and which solution should the next change start from. Patterns in the history, such
as a plateau or a breakthrough, are left to the model's situation assessment rather than
becoming control states, because real histories mix patterns and the right response often
combines intentions. We do without a forest of dormant directions with revive operations: the
log records every line that was tried, and the controller can return to a direction by
starting a new line toward it from the baseline. We also do without a composite stack of
mechanisms, because the mechanisms are already embodied in the Champion's code and the
controller needs only its signature. Further intentions, such as verifying, decomposing, or
comparing, stay instructions rather than becoming actions, and retrieval belongs to Generate.

\begin{table}[t]
\centering
\small
\setlength{\tabcolsep}{3pt}
\begin{tabular}{@{}>{\raggedright}p{0.30\columnwidth}>{\raggedright}p{0.31\columnwidth}>{\raggedright\arraybackslash}p{0.33\columnwidth}@{}}
\toprule
\textbf{Left out} & \textbf{Intended purpose} & \textbf{Replacement} \\
\midrule
Pattern--action table & Map history to next action & LLM decision over the log \\
Research forest, revive states & Return to old directions & The log plus a new line from the baseline \\
Composite mechanism stack & Track accumulated solution & Champion signature plus code \\
Multi-axis round plan & Express fine-grained intent & Goal-level instruction plus hard constraints \\
\bottomrule
\end{tabular}
\caption{Structures left out of the controller and what replaces them.}
\label{tab:removed}
\end{table}

\section{Implementation on an Existing AI-Scientist Stack}
\label{app:implementation}

\label{sec:implementation}

The controller is an orchestration-layer component rather than a replacement for the
execution stack. In a typical AI-scientist implementation, existing components can be reused
as in Table~\ref{tab:mapping}.

\begin{table}[t]
\centering
\small
\setlength{\tabcolsep}{4pt}
\begin{tabular}{@{}l>{\raggedright\arraybackslash}p{0.68\columnwidth}@{}}
\toprule
\textbf{Phase} & \textbf{Implementation role} \\
\midrule
Decide & New controller LLM node over the Champion and the log \\
Generate & Existing search planner, retriever, and idea generator \\
Run & Existing code runner and scheduler \\
Record & New log writer and Champion updater \\
Base copying & Existing parent-workspace copy mechanism, if available \\
\bottomrule
\end{tabular}
\caption{Mapping the loop's phases to common AI-scientist execution components.}
\label{tab:mapping}
\end{table}

The main replacement is the fixed iteration rule found in many stacks, such as ``round one is
diverse, later rounds improve,'' which gives way to a Decide step that reads the Champion and
the log.

\paragraph{What the design requires of an execution stack.}
\textsc{Decide} is agnostic to the orchestration framework and to the models: its interface is
the floor, the Champion, and the log in, and an action, a goal-level instruction, and hard
constraints out. Any execution stack that meets two conditions can host it. First, it must
turn a natural-language instruction into a runnable code change in a given workspace and
report a metric, which a modern coding agent already provides. Second, it must keep the
Champion's workspace, because extending and refining the Champion both start from a copy of
it. In our stack, \textsc{Run} invokes the Claude Code agent through the
\texttt{claude-agent-sdk}; the agent plans, writes, executes, reviews, and repairs the
candidate's code inside its own workspace with file and shell tools, and \textsc{Record}
parses the metric the workspace reports.

\paragraph{Failure handling.}
Candidate failures are not discarded. Their diagnoses enter the log and can steer later rounds
away from repeated dead ends. This matters in scientific code generation, where a failed
implementation may still reveal that a method is too brittle, too expensive, or mismatched to
the dataset.

\section{The Generate Agents and the Decide Prompt}
\label{app:generate}

This section details the three agents that carry out \textsc{Generate}
(Figure~\ref{fig:method}, Section~\ref{sec:method}) and gives the prompt of \textsc{Decide}.
All of them call \texttt{deepseek-v4-pro}.

\subsection{The Generate agents}

\paragraph{Search Planner.}
For both exploration actions, the Search Planner makes one LLM call per round. Its prompt holds
the task description, domain, and metric, an analysis of the baseline, the controller's
instruction, situation assessment, and rationale, the survey, and a ledger of tried directions.
The ledger has one entry per earlier round: its action, starting point, goal, and hard
constraints, every candidate with its mechanism signature, metric, and difference from its
starting point, and the round's lesson. The ledger is rendered from the log in every context
setting of Section~\ref{sec:ablation}, since that study varies only what \textsc{Decide} reads.
The planner returns six search plans. Each has a short paper-style title, which serves as the
retrieval query, keywords, a rationale, and a descriptive label (focused, cross-domain,
theoretical, or corrective) that does not change how the plan is retrieved.

\paragraph{Literature Retriever.}
Each plan is retrieved separately, three plans at a time. Its title goes to a retrieval
service that searches a pre-built arXiv index with dense embeddings
(\texttt{gte-large-en-v1.5}, top 120), reranks the hits with a cross-encoder
(\texttt{ms-marco-MiniLM-L6-v2}, top 60), and queries the arXiv API when fewer than 20 hits come
back from the index. It returns up to 30 papers; papers whose abstract is shorter than 50
characters are dropped. The LLM then ranks the papers in up to two passes. With 15 or more
papers, a first pass ranks all of them by relevance to the plan from their titles and short
summaries. The top 15 go on to a second pass, which ranks them by relevance and then by
usefulness for generating ideas (mechanism insight, transferable modules, implementation
clues) and is told not to rank by date, venue, citations, or author reputation. With 3 to 14
papers only the second pass runs, and with fewer than 3 no ranking is needed. PDFs are
downloaded and parsed in ranked order until three have been parsed, and the LLM writes a deep
read of each with four fields: an overview (problem, core idea, contributions), the methodology
(architecture, algorithm, key formulas), the results, and takeaways (transferable techniques,
limitations).

\paragraph{Idea Forge.}
For exploration, the Idea Forge works in two stages. It first drafts one concept per plan from
the overviews of that plan's papers, the task and metric, the instruction, and the hard
constraints; a concept names the idea, its motivation, and the papers it draws on. It then
develops each concept into a method with implementation details, from the methodology fields of
those papers and the starting code: the baseline's code for a new line, or the Champion's code
when the Champion is extended. An exploration round therefore yields up to six ideas, fewer when
a plan yields no parsed paper. The LLM, prompted as a strict reviewer, then scores each idea on
two axes in $[0,1]$. Feasibility is how likely its code is to run and yield a valid metric
without breaking the evaluation protocol, judged by the size of the change and the number of new
modules. Grounding is whether the idea builds on a technique from the round's retrieved papers
and whether its main effect targets the primary metric. The reviewer is told not to predict how
much an idea will improve the metric. An idea passes if both scores are at least $0.4$. If at
least $N$ ideas pass, the $N$ most feasible of them are run; otherwise the $N$ most feasible of
all ideas are run.

For refinement, the Idea Forge uses no papers. It reads the Champion's \texttt{experiment.py}
and drafts five concepts, each an incremental change to a named part of that code and each
distinct in mechanism from the others. They are developed on the Champion's code and scored
as above. Because refinement ideas cite no retrieved papers, they rarely pass the grounding gate,
so in refinement rounds the candidates are usually the $N$ most feasible of the five.

\subsection{The Decide prompt}
\label{app:decide-prompt}

\textsc{Decide} sends one system prompt and one user message per round and parses the reply into
seven fields. The action of Section~\ref{sec:decide} is the pair of \texttt{mode} and
\texttt{base}; \texttt{situation} is the assessment the model writes first, and
\texttt{stop\_reason} is filled only on a stop. The three boxes below give the system prompt, the user message template, and the
output fields. The templates keep the words exactly as used and change only the typography:
lines are rewrapped to the box, list markers are drawn as bullets, headings are coloured, and the
placeholders in braces, which the code fills in, are shaded. Every
\textsc{Decide} call logged in the runs reported in this paper used these two templates. The code
renders the Champion and the log into the user message; in the context settings of
Section~\ref{sec:ablation} it renders only what that setting allows. After parsing, the guardrails
described in Section~\ref{sec:decide} are enforced in code. The prompt's heading ``Three actions''
counts the actions that run experiments and lists \texttt{stop} with them.

\begin{promptbox}{Decide: system prompt}{Sys}
\input{sections/prompts/decide_system}
\end{promptbox}

\begin{promptbox}{Decide: user message template}{Usr}
\input{sections/prompts/decide_user}
\end{promptbox}

\begin{promptbox}{Decide: output fields}{Out}
\renewcommand{\arraystretch}{1.15}%
\begin{tabular}{@{}l@{\hspace{7pt}}>{\raggedright\arraybackslash}p{0.58\linewidth}@{}}
\textcolor{pbAccent}{\ttfamily\bfseries situation} & The model's assessment of the state, written first \\[1.5pt]
\textcolor{pbAccent}{\ttfamily\bfseries mode} & \PromptCode{explore}, \PromptCode{exploit}, or \PromptCode{stop} \\[1.5pt]
\textcolor{pbAccent}{\ttfamily\bfseries base} & \PromptCode{baseline} or the Champion; with \PromptCode{mode} it gives the action: new line, extend, or refine \\[1.5pt]
\textcolor{pbAccent}{\ttfamily\bfseries instruction} & The round's goal, at most 40 words \\[1.5pt]
\textcolor{pbAccent}{\ttfamily\bfseries hard\_constraints} & Zero to three non-negotiable, checkable requirements \\[1.5pt]
\textcolor{pbAccent}{\ttfamily\bfseries rationale} & Why, citing evidence from the log \\[1.5pt]
\textcolor{pbAccent}{\ttfamily\bfseries stop\_reason} & Filled only when \PromptCode{mode} is \PromptCode{stop} \\[1.5pt]
\end{tabular}
\end{promptbox}

\section{Re-running InternAgent with Our Models}
\label{app:internagent-rerun}

To separate the controller from the backbone, we re-ran InternAgent's released code (the
public NovelSeek repository, commit \texttt{fa8c3ee}) on the three benchmark tasks with the
models our controller uses: \texttt{deepseek-v4-pro} for every ideation, survey, review, and
ranking agent, and a Claude Code coding agent backed by \texttt{MiniMax-M3} for experiment
execution, the same backend as our \textsc{Run} step, on the same hardware as our runs. The
reported number is the best verified metric over the run, the quantity reported in
Table~\ref{tab:main-results}.

\section{Per-Round Trajectories of the Four Main Runs}
\label{app:trajectories}

The four tables below give the round-by-round record of the runs summarised in
Section~\ref{sec:evaluation}; Figure~\ref{fig:case} draws the third of them, the AutoCls2D run.
\#cand.\ is the number of candidates run in the round and \#fail the number without a valid
metric. A base other than the baseline is the Champion, written as its identifier
\texttt{iter\_$t$\_idea\_$k$}, where $k$ indexes the ideas generated in round $t$, of which
only the $N$ selected were run.

\begin{table}[t]
\centering
\small
\setlength{\tabcolsep}{3pt}
\begin{tabular}{@{}r l r r r@{}}
\toprule
R & Decide (mode/base) & \#cand. & \#fail & Best MAE $\downarrow$ \\
\midrule
1 & \emph{explore}/baseline & 3 & 0 & 0.4608 \\
2 & \emph{explore}/baseline & 3 & 0 & 0.4947 \\
3 & \emph{explore}/baseline & 3 & 1 & 0.5443 \\
4 & \emph{explore}/baseline & 3 & 0 & \textbf{0.4226} \\
5 & \emph{explore}/\texttt{iter\_4\_idea\_3} & 3 & 0 & \textbf{0.4193} \\
6 & \emph{explore}/\texttt{iter\_5\_idea\_1} & 3 & 1 & \textbf{0.4184} \\
7 & \emph{explore}/\texttt{iter\_6\_idea\_4} & 3 & 0 & \textbf{0.4172} \\
8 & \emph{exploit}/\texttt{iter\_7\_idea\_3} & 3 & 0 & \textbf{0.4172} \\
9 & \emph{stop} & --- & --- & --- \\
\bottomrule
\end{tabular}
\caption{Main run on AutoTSF (\texttt{20260731\_080205}); \textsc{Decide} reads the Champion and the full log. The controller sets the first Champion at R4 and improves it in each of R5--R8, reaching MAE $0.4172$ (R8 improves on R7 only in the fifth decimal, $0.41724 \to 0.41721$), and stops at R9 with budget left, after eight of ten budgeted rounds. Baseline MAE: 0.4382. Three candidates per round. \textbf{Bold} marks rounds that improved the Champion.}
\label{tab:main-run}
\end{table}

\begin{table}[t]
\centering
\small
\setlength{\tabcolsep}{3pt}
\begin{tabular}{@{}r l r r r@{}}
\toprule
R & Decide (mode/base) & \#cand. & \#fail & Best Acc $\uparrow$ \\
\midrule
1 & \emph{explore}/baseline & 3 & 1 & \textbf{94.61} \\
2 & \emph{exploit}/\texttt{iter\_1\_idea\_1} & 3 & 0 & 94.15 \\
3 & \emph{explore}/baseline & 3 & 0 & \textbf{94.95} \\
4 & \emph{exploit}/\texttt{iter\_3\_idea\_3} & 3 & 0 & 94.04 \\
5 & \emph{explore}/baseline & 3 & 0 & 93.81 \\
6 & \emph{explore}/\texttt{iter\_3\_idea\_3} & 3 & 0 & 94.27 \\
7 & \emph{explore}/baseline & 3 & 0 & 94.38 \\
8 & \emph{exploit}/\texttt{iter\_3\_idea\_3} & 3 & 0 & \textbf{95.07} \\
9 & \emph{exploit}/\texttt{iter\_8\_idea\_1} & 3 & 0 & 94.38 \\
10 & \emph{stop} & --- & --- & --- \\
\bottomrule
\end{tabular}
\caption{Main run on AutoClsSST (\texttt{20260803\_001220}), ten-round budget, three candidates per round; \textsc{Decide} reads the Champion and the full log. Accuracy (\%) on SST-2. The controller stops at R10, after nine executed rounds. Baseline accuracy: $91.63\%$. \textbf{Bold} marks rounds that improved the Champion.}
\label{tab:main-run-clssst}
\end{table}

\begin{table}[t]
\centering
\small
\setlength{\tabcolsep}{3pt}
\begin{tabular}{@{}r l r r r@{}}
\toprule
R & Decide (mode/base) & \#cand. & \#fail & Best Acc $\uparrow$ \\
\midrule
1 & \emph{explore}/baseline & 4 & 0 & \textbf{82.20} \\
2 & \emph{explore}/\texttt{iter\_1\_idea\_5} & 4 & 1 & \textbf{83.57} \\
3 & \emph{explore}/\texttt{iter\_2\_idea\_2} & 4 & 1 & \textbf{83.91} \\
4 & \emph{explore}/\texttt{iter\_3\_idea\_4} & 4 & 0 & 83.39 \\
5 & \emph{explore}/baseline & 4 & 0 & 81.62 \\
6 & \emph{explore}/baseline & 4 & 1 & 80.51 \\
7 & \emph{explore}/baseline & 4 & 1 & 82.05 \\
8 & \emph{exploit}/\texttt{iter\_3\_idea\_4} & 4 & 0 & \textbf{84.29} \\
9 & \emph{exploit}/\texttt{iter\_8\_idea\_3} & 4 & 0 & 84.24 \\
10 & \emph{exploit}/\texttt{iter\_8\_idea\_3} & 4 & 1 & \textbf{84.63} \\
\bottomrule
\end{tabular}
\caption{Main run on AutoCls2D (\texttt{20260923\_131120}), ten rounds, four candidates per round; \textsc{Decide} reads the Champion and the full log. Top-1 accuracy on CIFAR-100. The in-run baseline is a three-epoch smoke configuration ($49.23\%$) that serves only as the floor before the first Champion. \textbf{Bold} marks rounds that improved the Champion.}
\label{tab:main-run-cls2d}
\end{table}

\begin{table}[t]
\centering
\small
\setlength{\tabcolsep}{3pt}
\begin{tabular}{@{}r l r r r@{}}
\toprule
R & Decide (mode/base) & \#cand. & \#fail & Best Macro-F1 $\uparrow$ \\
\midrule
1 & \emph{explore}/baseline & 5 & 0 & \textbf{56.58} \\
2 & \emph{explore}/\texttt{iter\_1\_idea\_6} & 5 & 0 & \textbf{59.94} \\
3 & \emph{explore}/\texttt{iter\_2\_idea\_2} & 5 & 1 & 55.65 \\
4 & \emph{explore}/baseline & 5 & 5 & --- \\
5 & \emph{explore}/\texttt{iter\_2\_idea\_2} & 5 & 5 & --- \\
6 & \emph{explore}/baseline & 5 & 0 & 56.52 \\
7 & \emph{exploit}/\texttt{iter\_2\_idea\_2} & 5 & 0 & \textbf{60.19} \\
8 & \emph{exploit}/\texttt{iter\_7\_idea\_2} & 5 & 0 & \textbf{60.19} \\
9 & \emph{exploit}/\texttt{iter\_8\_idea\_4} & 4 & 1 & 60.13 \\
10 & \emph{stop} & --- & --- & --- \\
\bottomrule
\end{tabular}
\caption{Main run on AutoDepDAIC (\texttt{20260814\_164352}), ten-round budget,
five candidates per round (four in R9); \textsc{Decide} reads the Champion and the full log. Nine rounds
executed, after which the controller stopped on convergence. R8 improves on R7 only in the
third decimal ($60.188 \to 60.191$). Values are
macro-F1~$\times 100$ under pooled 188-subject subject-stratified 5-fold CV over five
seeds; baseline floor: $56.10$. \textbf{Bold} marks rounds that improved the Champion.}
\label{tab:main-run-depdaic}
\end{table}

\section{AutoDepDAIC Evaluation Protocol}
\label{app:depdaic}

DAIC-WOZ has a documented history of inflated results because the literature convention
selects models on the same development split it reports.
HAREN-CTC~\citep{li2026haren} quantifies the cost of that convention on its own model: its
macro-F1 drops from $0.81$ under dev-selection to $0.576$ under a leakage-free protocol. We
adopt the latter. All 188 usable subjects are pooled and evaluated by subject-stratified
5-fold cross-validation repeated over five seeds, and every training, hyper-parameter, and
threshold choice is made strictly inside each fold's training part, so the evaluated fold is
never visible to the model that scores it. The primary metric is macro-F1, chosen because
the depressed class is a minority: the all-positive degenerate predictor scores only
$\approx 0.23$ and is therefore correctly suppressed, whereas the depressed-class F1 alone
would reward it. AUC serves as a threshold-free secondary check that distinguishes a
genuinely better ranking from a threshold that happened to move. The baseline floor is a
published recipe assembled from open components---frozen wav2vec2-base layer-12 embeddings
with a logistic probe---at macro-F1 $0.5610$ / AUC $0.6314$.

\paragraph{The discovered model.}
The Champion (R8) applies depression-contrastive speaker-direction removal followed by learned
attention over thirteen cached wav2vec2 layers, concentrating $0.81$ of its attention mass on
transformer layer~6. The fresh-seed re-run uses seeds 100--104, none of which took part in any
selection during the run.

\section{When Each Setting First Beats the Baseline}
\label{app:ablation-rounds}

Table~\ref{tab:ablation-ladder} reports the best verified result of each context setting. A
different quantity is the first round in which a setting beats the baseline. On AutoTSF the
runs of Table~\ref{tab:ablation-ladder} first beat the $0.4382$ floor at R1 (1-round history),
R2 (no history), and R4 (full history). The two settings that cross first finish behind the
full controller ($0.4244$ and $0.4299$), while the full controller crosses later and finishes first
($0.4172$). Beating the baseline once is easy, and a single good draw achieves it; what
separates the settings is how far the loop gets from there. We therefore treat the
first-crossing round as a diagnostic rather than a measure of controller quality. On AutoCls2D
the in-repository floor is a three-epoch smoke configuration, which every setting beats in its
first round.

%% file: sections/prompts/decide_system.tex
\PromptLine{0}{You are the strategy controller of an autonomous research loop. Every round you read the current best solution (champion) and the full research trajectory, then decide whether to refine the champion or switch to a new direction. The system will run N experiments accordingly, update the champion, and append to the trajectory.}
\PromptGap
\PromptLine{0}{Your single objective: on task ``\PromptSlot{task\_goal}'', push the champion's primary metric ``\PromptSlot{primary\_metric} (\PromptSlot{metric\_direction})'' as far as possible.}
\PromptGap
\PromptLine{0}{\PromptHead{[Three actions]} (which one applies is governed by the Budget-phase policy below)}\PromptKeep
\PromptItem{0}{1}{exploit + champion: incremental refinement on the champion's accumulated code (tuning / micro-mechanism adjustments). Do NOT introduce new external knowledge.}
\PromptItem{0}{1}{explore + champion: keep the champion's accumulated code, retrieve external literature, and stack a NEW mechanism on top of it.}
\PromptItem{0}{1}{explore + baseline: start from a clean baseline, retrieve external knowledge, build a new direction, and abandon the current lineage.}
\PromptItem{0}{1}{stop: either the loop has converged (even full direction switches stop improving) or the budget is exhausted. Deliver the champion.}
\PromptGap
\PromptLine{0}{\PromptHead{[Budget-phase policy (core)]}}\PromptKeep
\PromptLine{0}{You have \PromptSlot{rounds\_total} total rounds, \PromptSlot{rounds\_done} done, \PromptSlot{rounds\_left} left.}
\PromptGap
\PromptItem{0}{1}{Mid-phase (rounds\_left > reserve \textasciitilde{}= max(1, ceil(0.3 * \PromptSlot{rounds\_total}))):}
\PromptLine{1}{Choose explore or exploit.}
\PromptItem{2}{2}{explore+champion (default compounding engine) -- keep the best code, retrieve literature, and stack a new mechanism.}
\PromptItem{2}{2}{explore+baseline -- when champion-explore keeps stalling (no metric progress), abandon the current lineage and look for upside elsewhere.}
\PromptItem{2}{2}{exploit+champion -- when you can name the exact knobs / hyperparameters / training-strategy changes to try on the champion (no retrieval, no new module), refine the champion's real code directly. Exploit ALWAYS runs on the champion, NEVER on baseline.}
\PromptGap
\PromptItem{0}{1}{End-phase (rounds\_left <= reserve): switch to closing. Choose ONE of:}
\PromptItem{2}{2}{exploit+champion -- tune / refine the best until the gain is locked in;}
\PromptItem{2}{2}{stop -- deliver the champion.}
\PromptLine{1}{Exploit requires a REAL champion (strictly better than baseline). If there is NO champion yet even in end-phase, do NOT exploit -- keep explore+baseline.}
\PromptGap
\PromptLine{0}{\PromptHead{[Stall signal (decides base switching)]}}\PromptKeep
\PromptLine{0}{The user message tells you how many consecutive champion-explore rounds showed no progress. Default base is champion.}
\PromptItem{0}{1}{After roughly 2 consecutive rounds with no metric progress, switch to explore+baseline for a horizontal direction change.}
\PromptItem{0}{1}{Exception: if a candidate's ``realized'' (post-hoc) summary shows a structural advance that has not yet been reflected in the metric, you may give the champion one more round -- but justify it in \PromptCode{rationale}.}
\PromptItem{0}{1}{Do not thrash: only switch base when trajectory / realized evidence supports it.}
\PromptItem{0}{1}{base is ONLY \{baseline, champion\}. Do NOT revive an arbitrary historical candidate as base.}
\PromptItem{0}{1}{You own the direction (champion vs baseline): default to building on the champion, but when trajectory / realized evidence shows the lineage has stalled or a fresh direction has more upside, switch to explore+baseline on your own judgment. A hard backstop only forces the cross-jump once the explore-on-champion stall reaches the limit.}
\PromptGap
\PromptLine{0}{\PromptHead{[How to write the instruction -- GOAL-ONLY, at most 40 words]}}\PromptKeep
\PromptItem{0}{1}{The instruction is the round's RESEARCH GOAL in ONE sentence (<= 40 words):}
\PromptLine{1}{name the bottleneck to attack and roughly the research area / target level. Do NOT pack implementation details, retrieval keywords, or multi-item recipes into it -- SearchPlanner expands the goal into diverse queries, IdeaForge proposes the concrete mechanisms, and any NON-NEGOTIABLE item goes into \PromptCode{hard\_constraints}, not into instruction prose.}
\PromptLine{1}{OK (goal/bottleneck): ``champion is limited by cross-channel dependency modeling on ETTh1; explore diverse strong approaches to cross-variable modeling, aiming toward the \textasciitilde{}0.42 frontier level.''}
\PromptLine{1}{NOT OK (too narrow): ``anchor ModernTCN and add a channel-mixing gate.''}
\PromptLine{1}{NOT OK (too vague): ``find a better method.''}
\PromptLine{1}{NOT OK (recipe list): ``(1) raise max\_seq\_len to 128, (2) add LLRD, (3) tune LR.''}
\PromptItem{0}{1}{For EXPLOIT: still <= 40 words but concrete about WHAT on the champion you are changing (no retrieval happens in exploit); when exact knob values are non-negotiable, put them into hard\_constraints.}
\PromptGap
\PromptLine{0}{\PromptHead{[hard\_constraints -- the non-negotiable channel]}}\PromptKeep
\PromptItem{0}{1}{0-3 items, each concrete and checkable post-hoc; examples are task-dependent (e.g. ``set max\_seq\_len=128'', ``keep the baseline's normalization scheme'', ``do not change the evaluation split''). Constraints reach EVERY candidate's generation and implementation VERBATIM, bypassing paraphrase; after the round, compliance is checked and recorded in the trajectory.}
\PromptItem{0}{1}{Use them ONLY for genuinely non-negotiable requirements; leave the list empty by default. Candidates must implement constraints as-is and innovate ON TOP of them -- creativity rules do not apply to the constraints themselves.}
\PromptGap
\PromptItem{0}{1}{The instruction MUST be self-consistent with this round's mode/base:}
\PromptItem{2}{2}{exploit: only talk about incremental changes on the champion. Do NOT mention reproducing or retrieving a new SOTA.}
\PromptItem{2}{2}{explore+baseline: a fresh direction/goal that abandons the champion lineage (still goal-level, not one pinned method).}
\PromptItem{2}{2}{explore+champion: keep the champion as base and aim at the round's goal/bottleneck on top of it; IdeaForge picks the specific mechanism.}
\PromptLine{1}{A decision whose frame and instruction conflict is invalid.}
\PromptGap
\PromptLine{0}{\PromptHead{[Use the frontier as a map of strong targets -- anchor at the IDEA level]}}\PromptKeep
\PromptLine{0}{The user message includes a ``Frontier Research Landscape'' (validated research lines + representative results under the same setting). Use it to set the round's TARGET, NOT to pin one method:}
\PromptItem{0}{1}{An explore instruction names the target LEVEL / AREA to aim at (e.g. ``reach the \textasciitilde{}0.42 strong-line level by improving cross-channel modeling''), NOT one specific method + one specific improvement. Grounding to a validated technique happens PER-IDEA in IdeaForge -- do NOT pre-collapse it to a single named method in the directive.}
\PromptItem{0}{1}{Reliability lives at the IDEA level: each idea must ground in a validated technique and be ONE clean mechanism (no stacks of unproven modules -- that historically underperforms). This keeps breadth (diverse ideas) without unanchored ad-hoc piles.}
\PromptItem{0}{1}{At cold start (no champion), the directive sets the target/area; IdeaForge then explores diverse strong directions toward it.}
\PromptItem{0}{1}{Keep diversity: within the frontier map, choose a direction that is well-grounded yet under-explored in the current trajectory, and avoid repeating already-failed candidates.}
\PromptItem{0}{1}{Honor any prerequisite components the landscape flags as standard practice; ignoring established prerequisites usually wastes a round.}
\PromptItem{0}{1}{The frontier is a DIRECTION reference, NOT a reachability claim. A published metric (even if better than your champion) is aspirational, not free headroom. A REAL, in-run validated champion OUTRANKS an unreached published number -- do NOT abandon the champion just because the landscape cites a stronger figure.}
\PromptItem{0}{1}{If a research line has already FAILED to run / reproduce in this run's trajectory (see the briefing), treat that as NEGATIVE evidence about its feasibility here -- do NOT read repeated failures as ``under-explored, worth chasing again.''}
\PromptItem{0}{1}{LOCAL EVIDENCE OVERRIDES the frontier: when this run's own trajectory (a round's lesson line, a refuted hypothesis) contradicts a frontier expectation, trust the local measurement, and state the conflict explicitly in \PromptCode{situation} instead of re-issuing the frontier narrative.}
\PromptGap
\PromptLine{0}{\PromptHead{[Output]}}\PromptKeep
\PromptLine{0}{Strictly follow the DecisionOutput schema: first write \PromptCode{situation} (assess the state), then provide \PromptCode{mode} / \PromptCode{base} / \PromptCode{instruction} (goal-only, <= 40 words) / \PromptCode{hard\_constraints} (usually empty) / \PromptCode{rationale}. Default base = champion (mid-phase explore stacks on the best; end-phase exploit polishes the best). base is ONLY \{baseline, champion\}. Only switch to explore+baseline when the champion has stalled and a horizontal direction change is warranted. The directive (what to actually research / change) is always yours to decide; just keep it self-consistent with mode/base.}

%% file: sections/prompts/decide_user.tex
\PromptLine{0}{\PromptHead{\# Task goal}}\PromptKeep
\PromptLine{0}{\PromptSlot{task\_goal}; primary metric \PromptSlot{primary\_metric} (\PromptSlot{metric\_direction})}
\PromptGap
\PromptLine{0}{\PromptHead{\# Current situation}}\PromptKeep
\PromptItem{0}{1}{Baseline floor: primary metric \PromptSlot{baseline\_metric}}
\PromptItem{0}{1}{Champion (best so far): \PromptSlot{champion\_desc}}
\PromptItem{0}{1}{Stall signal: \PromptSlot{stall\_hint}}
\PromptItem{0}{1}{Budget remaining: \PromptSlot{rounds\_left} rounds left}
\PromptGap
\PromptLine{0}{\PromptHead{\# Research trajectory (decision briefing: current best + full causal backbone + recent-K detail)}}\PromptKeep
\PromptLine{0}{\PromptSlot{trajectory\_render}}
\PromptGap
\PromptLine{0}{\PromptHead{\# Frontier Research Landscape (validated research lines + representative results under the same setting)}}\PromptKeep
\PromptLine{0}{\PromptSlot{frontier}}
\PromptGap
\PromptLine{0}{\PromptHead{\# Your decision}}\PromptKeep
\PromptLine{0}{Read the champion, the full trajectory, and the frontier landscape above, then output this round's decision following DecisionOutput.}
\PromptGap
\PromptLine{0}{Keep the instruction at GOAL level (<= 40 words) and within the frontier landscape's validated directions; put any non-negotiable requirement into \PromptCode{hard\_constraints} instead of instruction prose. When trajectory evidence contradicts a frontier expectation, trust the local evidence and say so in \PromptCode{situation}.}